\pdfoutput=1

\documentclass[11pt]{article}

\usepackage[final]{acl}

\usepackage{times}
\usepackage{latexsym}

\usepackage[T1]{fontenc}

\usepackage[utf8]{inputenc}

\usepackage{microtype}

\usepackage{inconsolata}

\usepackage{amsmath}
\usepackage{graphicx}
\usepackage{algorithm}
\usepackage{subcaption}
\usepackage{algorithmic}
\usepackage{booktabs}
\usepackage{multirow}
\usepackage{makecell}
\usepackage{tcolorbox}
\tcbuselibrary{listings}
\title{Contradiction Detection in RAG Systems: Evaluating LLMs as Context Validators for Improved Information Consistency}

\author{Vignesh Gokul \\
  \texttt{vgokulgv@amazon.com} \\\And
  Srikanth Tenneti \\
  \texttt{stenneti@amazon.com} \\\And
  Alwarappan Nakkiran \\
  \texttt{nakkiran@amazon.com}
  }

\begin{document}
\maketitle
\begin{abstract}
Retrieval Augmented Generation (RAG) systems have emerged as a powerful method for enhancing large language models (LLMs) with up-to-date information. However, the retrieval step in RAG can sometimes surface documents containing contradictory information, particularly in rapidly evolving domains such as news. These contradictions can significantly impact the performance of LLMs, leading to inconsistent or erroneous outputs. This study addresses this critical challenge in two ways. First, we present a novel data generation framework to simulate different types of contradictions that may occur in the retrieval stage of a RAG system. Second, we evaluate the robustness of different LLMs in performing as context validators, assessing their ability to detect contradictory information within retrieved document sets. Our experimental results reveal that context validation remains a challenging task even for state-of-the-art LLMs, with performance varying significantly across different types of contradictions. While larger models generally perform better at contradiction detection, the effectiveness of different prompting strategies varies across tasks and model architectures. We find that chain-of-thought prompting shows notable improvements for some models but may hinder performance in others, highlighting the complexity of the task and the need for more robust approaches to context validation in RAG systems.
\end{abstract}
\section{Introduction}
Large language models (LLMs) \cite{brown2020language} have become ubiquitous in a wide range of natural language processing applications, from chatbots to text generation systems. However, a key limitation of using LLMs is that their knowledge is static, reflecting only the information available during the training process. As a result, these models may lack the latest up-to-date facts and information needed for real-world tasks. To address this challenge, researchers have explored techniques like Retrieval Augmented Generation (RAG) \cite{lewis2020retrieval}, where relevant documents are dynamically retrieved and provided as context to the LLM. While this approach can help improve the model's knowledge, it introduces new problems related to contextual conflicts. Specifically, there are two main types of conflicts that can arise: 1) Context-memory conflict: cases where the retrieved context contradicts the parametric knowledge learned by the LLM during training, 2) Context-context conflict: situations where the retrieved contextual information itself contains contradictory statements. This work focuses on the latter issue of conflicts in the retrieved documents. Effectively detecting and resolving such conflicts is crucial for ensuring the reliability and consistency of LLM applications that rely on dynamic retrieval of external information. 

Detecting contradictions in text is a challenging task for several reasons. First, there is a scarcity of large-scale datasets specifically focused on contradiction detection. This lack of comprehensive data makes it difficult to train and evaluate systems effectively. Second, contradictions can be quite subtle and complex. While some contradictions are straightforward, such as conflicting numbers, others involve intricate logical inconsistencies that are not easily spotted. These nuances make it hard for models to reliably identify contradictions. In fact, psychological studies \cite{graesser1993anomalous, otero1992failures} have shown that even humans struggle with this task. Furthermore, recent research \cite{li2023contradoc} has revealed that advanced language models like GPT-4, GPT-3.5, and LLaMA-3 perform only slightly better than random guessing when it comes to detecting contradictions. This highlights the significant challenge that contradiction detection poses. 

Our primary objective in this study is to evaluate the effectiveness of  LLMs as context validators in RAG systems. The context validator is responsible for analyzing the retrieved context (set of documents) for contradictory information. Previous studies, such as \cite{9671319} and \cite{li2023contradoc}, have explored contradiction generation using Wikipedia templates and LLMs respectively. \cite{jiayang2024econ} evaluated LLMs' ability to detect contradictions in context pairs. However, these approaches do not fully address the complexities of the retrieval step in RAG systems. In a RAG-based system, multiple pieces of context are retrieved simultaneously, making it impractical to evaluate all possible pairs for contradictions. For instance, with just 20 retrieved documents, examining all 190 possible pairs for conflicts becomes unfeasible, given latency and cost considerations of practical RAG based systems. In this work we propose a novel framework for synthetic dataset generation that simulates various types of contradictions. 
 
Contradictions within retrieved document sets can manifest in subtle and nuanced ways, presenting significant challenges for RAG systems. In this study, we investigate three distinct types of contradictions that can occur in retrieved documents: 1) Self-contradictory documents, where a single document contains internally inconsistent information; 2) Contradicting document pairs, where two documents present conflicting information on the same topic; and 3) Conditional contradictions, involving a triplet of documents where the information in one document creates a contradiction between the other two. Examples of these types of contradictions are shown in Figure 1. Then, we design three tasks for context validation: detecting if any type of contradiction is present in the retrieved documents, predicting the type of contradiction present and finding the documents that are contradicting.  

\begin{figure*}[t]
    \centering
    \includegraphics[width = \textwidth]{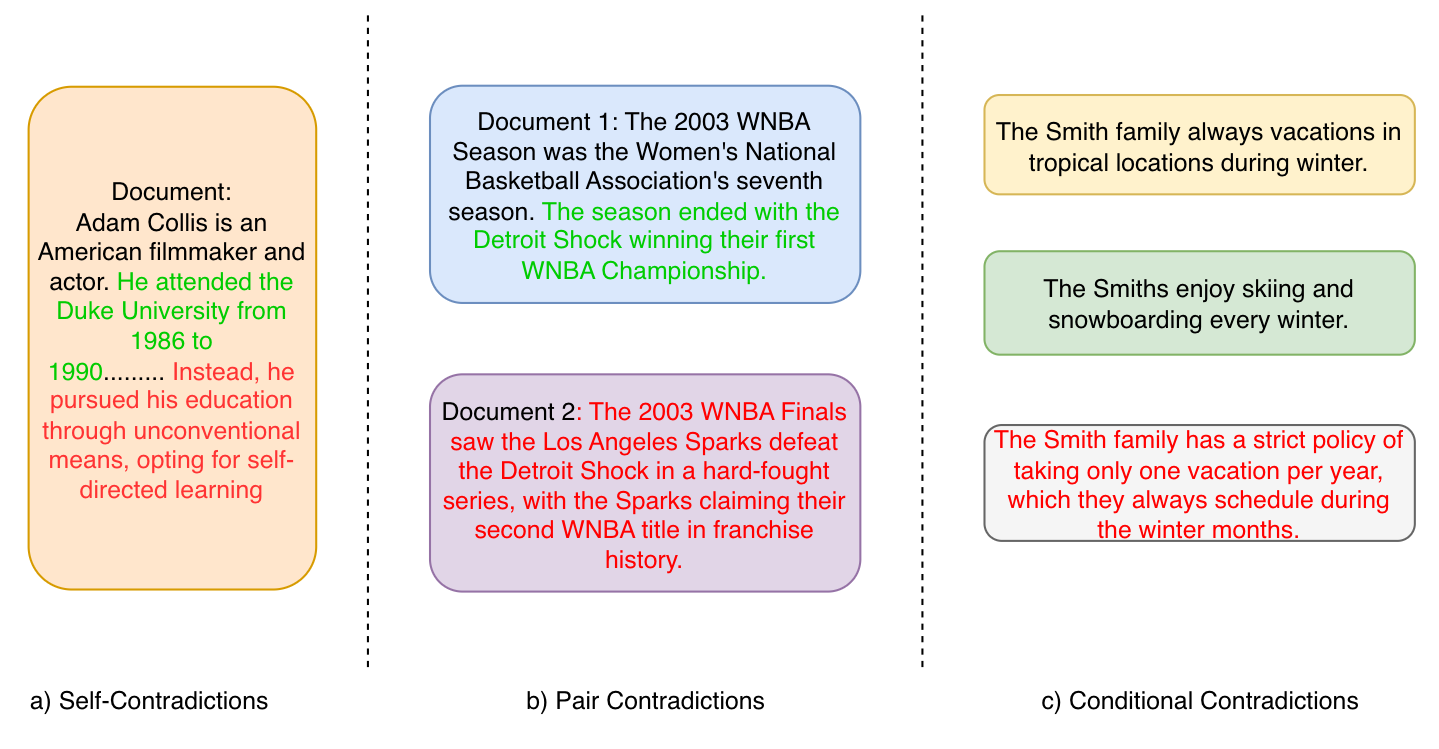}
    \caption{Different types of contradictions in the retrieved documents.}
    \label{fig:enter-label}
\end{figure*}

Our contributions can be summarized as follows:
\begin{itemize}
    \item  We introduce a novel synthetic data generation framework  that simulates diverse contradiction types in documents retrieved during the RAG process. This framework includes generating self-contradictory documents, pairwise contradictions, and conditional contradictions, providing a comprehensive testing dataset for evaluation purposes.
    \item We investigate the robustness of LLMs and prompting strategies in acting as a context validator in RAG systems: detecting conflicts in the retrieved documents (conflict detection), detecting the type of conflict (conflict type prediction) and identifying the contradicting documents (conflict segmentation).
    \item Our ablation studies provide empirical evidence and insights on the type of contradictions that are hard to detect by current state-of-the art LLMs.
\end{itemize}


\section{Related Work}
\textbf{Contradiction Detection:} Contradiction detection aims to classify if there are contradicting sentences in textual documents. Early research \cite{alamri2015automatic, badache2018predicting, lendvai2016monolingual} in this topic approached this problem as a supervised classification problem on a pair of sentences, i.e whether two sentences are contradicting or not. These works use linguistic features \cite{alamri2015automatic}, part-of-speech parsing \cite{badache2018predicting} or textual similarity features \cite{lendvai2016monolingual} to classify a pair of sentences. Contradiction detection could also be thought as a sub-class of Natural Language Inference (NLI). In NLI, the task is to label pairs of text as either neutral, entailment or contradictory. There have been numerous NLI works \cite{chen2016enhanced, mirakyan2018natural, parikh2016decomposable, rocktaschel2015reasoning}, with recent advances in transformer models \cite{vaswani2017attention} demonstrating strong capabilities for NLI \cite{devlin2018bert, liu2019roberta}. \cite{li2023contradoc} experimented with LLMs for contradiction detection. Their research proved that many LLMs struggle with the task of identifying conflicts in text data. However, existing literature does not explore how LLMs perform in detecting contradictions across multiple documents. Often, in retrieval based systems, multiple documents are retrieved. The above works focus on either 1 or 2 documents, while we analyze LLM to detect conflicts across many documents.

\textbf{Datasets for Contradiction Detection:}
\cite{9671319} propose WikiContradiction dataset by using Wikipedia templates to alter factual entities in statements to create contradictions. \cite{li2023contradoc} propose a LLM based generation of contradictions. To maintain document fluency while introducing contradiction, they use global fluency (perplexity based measure) and local fluency measures (BERT based score) to validate the contextual coherence of the modified sentences. Recently \cite{jiayang2024econ} proposed a data generation method to generate pairs of evidences that have contradictions. 

\section{LLMs as Context Validators} \label{sec:llmcv}
We formally define the problem of context validation as consisting of the following tasks: Given $N$ documents $D = \{d_1, d_2, \ldots, d_N\}$, a context validator is a function $f(D)$ such that:

\begin{equation}
f(D) = 
\begin{cases}
0|1 & \begin{tabular}{l} (conflict \\ detection) \end{tabular} \\[8pt]
t \in \mathcal{T} & \begin{tabular}{l} (conflict type \\ classification) \end{tabular} \\[8pt]
\{d_i, \ldots, d_m\} & \begin{tabular}{l} (conflicting context \\ segmentation) \end{tabular}
\end{cases}
\end{equation}

where, $0|1$ represents the binary output of conflict detection (0 for no conflict, 1 for conflict detected, $t \in \mathcal{T}$ is the classified type of conflict, $\{d_i, \ldots, d_m\}$ is the subset of documents containing conflicting contexts, where $1 \leq i \leq m \leq N$

Each document $d_i \in D$ is defined as an ordered set of $k_i$ statements, $d_i = \{s_1, s_2, \ldots, s_{k_i}\}$. In this work, we consider three types of contradictions in documents:

\begin{enumerate}
    \item \textbf{Self-contradictions:} Let $d_i \in D$. A self-contradiction occurs when $\exists s_p, s_q \in d_i$ such that $s_p$ contradicts $s_q$, where $p \neq q$.
    
    \item \textbf{Pair contradictions:} Let $d_i, d_j \in D$, where $i \neq j$. A pair contradiction occurs when $\exists s_p \in d_i, s_q \in d_j$ such that $s_p$ contradicts $s_q$.
    
    \item \textbf{Conditional contradictions:} Let $d_i, d_j, d_k \in D$, where $i \neq j \neq k$. A conditional contradiction occurs when $\exists s_p \in d_i, s_q \in d_j, s_r \in d_k$ such that:
    \begin{itemize}
        \item $s_p$ does not contradict $s_q$
        \item $s_p$ does not contradict $s_r$
        \item $s_r \implies (s_p \oplus s_q)$, the presence of $s_r$ implies that $s_p$ and $s_q$ are mutually exclusive
    \end{itemize}
\end{enumerate}


It is important to note that although these contradiction types involve one, two, or three documents respectively, the context validator function $f$ operates on the entire set of documents $D = \{d_1, \ldots, d_N\}$. This approach is crucial for several reasons. Firstly, it ensures computational efficiency. Inspecting all types of contradictions by examining documents in isolation, pairs, or triples would require $O(N)$, $O(N^2)$, and $O(N^3)$ LLM calls respectively, where $N = |D|$. Specifically, self-contradictions would require $N$ calls, pair contradictions would need $\binom{N}{2} = \frac{N(N-1)}{2}$ calls, and conditional contradictions would necessitate $\binom{N}{3} = \frac{N(N-1)(N-2)}{6}$ calls. This approach would significantly increase both the computational cost and latency of the LLM system. The design of $f$ as a function operating on the power set of $D$ (that is, $f: \mathcal{P}(D) \rightarrow \{0,1\} \times \mathcal{T} \times \mathcal{P}(D)$) addresses the limitations of conventional conflict detection methods. Furthermore, traditional approaches, such as Natural Language Inference (NLI) models, typically process only two texts at a time. This limitation makes them inadequate as context validators, particularly for identifying self-contradictions and conditional contradictions, which require analysis of one and three documents, respectively. 

In the following sections, we describe the data generation methods for each conflict type. Subsequently, we present the results of our evaluations on the synthetic data.

\textbf{Self-Contradictory Documents:} To generate synthetic self-contradictory documents, we sample a document $d_i = {s_1, s_2, \ldots, s_m}$ from $D$, where each $s_j$ represents a statement. First, we use a LLM to extract a sentence from the text, denoted as $s_i = \text{ChooseStatement}(d_i, \text{importance})$. The 'importance' parameter allows us to select either the most salient or least significant statement. Once a sentence has been extracted, we generate a contradicting statement $s_i' = \text{ContradictStatement}(s_i)$. To make detection more challenging, we then use an LLM to generate a paragraph incorporating the contradictory statement: $c_i' = \text{ContextGenerate}(s_i', \text{length})$. Here, we experiment with different 'length' values to vary the complexity and subtlety of the contradiction. The final step involves augmenting the original document $d_i$ with the generated contradictory context $c_i'$, resulting in a self-contradictory document $d_i' = d_i \cup {c_i'}$.

\textbf{Pair Contradictions:}
In pair contradictions, our objective is to induce contradictions across multiple documents. We follow a procedure similar to that used for generating self-contradictory documents, but with modifications to span multiple documents.
The conflicting context $c_j'$ is inserted into $D$, resulting in an updated set $D'$. We experiment with two configurations for the insertion: near and far. These configurations determine the indices of the contradicting documents in the document list. A 

\textbf{Conditional Contradictions:} To generate conditional contradictions, we start by sampling a document $d_i$ from our set D. We then extract the first sentence s from $d_i$ to serve as our "topic". Using an LLM, we generate three new documents on this topic: $d1'$, $d2'$, and $d3'$ = GenerateConditionalDocs($s$). These documents are generated with specific constraints: $d1'$ and $d2'$ should not contradict each other, $d3'$ should not directly contradict either $d1'$ or $d2'$, but the information in $d3'$ should make $d1'$ and $d2'$ mutually exclusive. This means that both $d1'$ and $d2'$ cannot be simultaneously true. We experiment with two configurations for inserting these documents into our set $D$ to get $D'$. In the contiguous setting, we keep the three documents near each other when inserting into $D$. In the separate setting, we spread the documents randomly across $D$.

Algorithm 1 shows the overall method for generating each conflict type. The prompts for each function in the algorithm are provided in the Appendix A.1.

\begin{algorithm}
\caption{Generate Synthetic Contradictions}
\begin{algorithmic}[1]
\REQUIRE Set $D$, parameters $\alpha$, $\lambda$
\ENSURE Set $D'$ with contradictions

\STATE \textbf{Function} GenSelfContrad($d_i$, $\alpha$, $\lambda$):
    \STATE \hspace{\algorithmicindent} $s_i \gets \text{ChooseStmt}(d_i, \alpha)$
    \STATE \hspace{\algorithmicindent} $s_i' \gets \text{Contradict}(s_i)$
    \STATE \hspace{\algorithmicindent} $c_i' \gets \text{GenContext}(s_i', \lambda)$
    \STATE \hspace{\algorithmicindent} $d_i' \gets d_i \cup \{c_i'\}$
    \STATE \hspace{\algorithmicindent} $D' \gets D - \{ d_i\} + \{d_i'\}$

\STATE \textbf{Function} GenPairContrad($d_i$, $\alpha$, $\lambda$, cfg):
    \STATE \hspace{\algorithmicindent} $s_i \gets \text{ChooseStmt}(d_i, \alpha)$
    \STATE \hspace{\algorithmicindent} $s_j' \gets \text{Contradict}(s_i)$
    \STATE \hspace{\algorithmicindent} $c_j' \gets \text{GenContext}(s_j', \lambda)$
    \STATE \hspace{\algorithmicindent} $D' \gets \text{Insert}(c_j', D, \text{cfg})$

\STATE \textbf{Function} GenCondContrad($d_i$, cfg):
    \STATE \hspace{\algorithmicindent} $s \gets \text{GetFirst}(d_i)$
    \STATE \hspace{\algorithmicindent} $d_1', d_2', d_3' \gets \text{GenCond}(s)$
    \STATE \hspace{\algorithmicindent} $D' \gets \text{Insert}(d_{1,2,3}', D, \text{cfg})$
\RETURN $D'$
\end{algorithmic}
\end{algorithm}

\begin{table*}[h]
\centering
\begin{tabular}{|p{2.5cm}|p{12cm}|}
\hline
\textbf{Conflict Type} & \textbf{Example} \\
\hline
Self-contradiction & \textbf{Document:} "Low pressure receptors are baroreceptors located in the venae cavae and the pulmonary arteries, and in the atria. High pressure receptors, rather than low pressure receptors, are baroreceptors located in the venae cavae and the pulmonary arteries, and in the atria.These baroreceptors monitor changes in blood pressure and relay this information to the cardiovascular control centers in the medulla oblongata of the brain...." \\
\hline
Pair contradiction & \textbf{Document 1:} "Apple Remote Desktop (ARD) is a Macintosh application produced by Apple Inc., first released on March 14, 2002, that replaced a similar product called "Apple Network Assistant"..."
\newline \textbf{Document 2:} "Apple Remote Desktop (ARD) is not a Macintosh application produced by Apple Inc., nor did it replace a similar product called "Apple Network Assistant".Apple Remote Desktop (ARD) is a software application developed by Apple Inc. that allows users to remotely control and manage other computers over a network..." \\
\hline
Conditional contradiction & \textbf{Document 1:} "David C is a passionate artist who creates captivating abstract paintings using a unique blend of techniques. His works have been featured in several prestigious art galleries and exhibitions."
\newline \textbf{Document 2:} "David C is a passionate artist who creates captivating abstract paintings using a unique blend of techniques. His works have been featured in several prestigious art galleries and exhibitions."
\newline \textbf{Document 3:} "David C dedicates his entire professional life to his work, devoting all his time and energy to a single pursuit, leaving no room for other significant commitments or interests." \\
\hline
\end{tabular}
\caption{Examples of Different Types of Textual Conflicts}
\label{tab:conflicts}
\end{table*}
\section{Data Analysis}
We construct our synthetic dataset using documents from HotpotQA \cite{yang2018hotpotqa}, a dataset known for its multi-hop reasoning requirements and diverse document content. Using Claude-3 Sonnet as our generation model, we created a dataset of 1,867 samples with varying types of contradictions. As shown in Appendix A.3 Table \ref{tab:dataset_stats}, we maintain a balanced distribution across different contradiction types, with 37.49\% containing no contradictions, serving as negative samples. Among the contradictory samples, self-contradictions comprise 26.30\%, followed by pair contradictions (19.07\%) and conditional contradictions (17.14\%).


To validate the quality of our synthetic dataset, we conducted a human evaluation study on 140 randomly sampled examples (50 each for self-contradictions and pair-contradictions, and 40 for conditional contradictions). Two expert annotators independently evaluated these documents for the presence of contradictions. To focus only on the quality of generated contradictions, only documents containing conflicts were presented to the annotators. We observed an overall inter-annotator agreement rate of 74\%. For conditional contradictions, annotators identified conflicts in only 17 of 40 examples, while for pair and self-contradictions, they marked 84 samples as contraditcions. Analyzing the remaining 16 samples, we found them to be contradictory as well (samples provided in Appendix). These results highlight two significant insights: our generation approach successfully creates subtle and nuanced contradictions that can escape initial detection, and contradiction detection poses significant challenges even for human experts. These findings align with previous studies on human performance in contradiction detection tasks \cite{graesser1993anomalous, otero1992failures} and highlight the challenging nature of our dataset.

\begin{table}[htbp]
\centering
\caption{Distribution of contradiction types}
\label{tab:dataset_stats}
\begin{tabular}{lrr}
\toprule
Type & Count & Pct (\%) \\
\midrule
None & 700 & 37.49 \\
Self & 491 & 26.30 \\
Pair & 356 & 19.07 \\
Cond. & 320 & 17.14 \\
\midrule
Total & 1,867 & 100.00 \\
\bottomrule
\end{tabular}
\end{table}
\section{Evaluation Setup}
We design three evaluation tasks: conflict detection, conflit type prediction and conflicting context segmentation. 
\subsection{Conflict Detection:} We ask the model to identify if there are any contradictions in the provided set of documents. We formalize this as a binary classification task: the model is tasked to answer "yes" or "no". For evaluation, we use the classification metrics: accuracy, precision, recall and F1 score.
\begin{table*}[t]
\centering
\caption{Performance of Various Models and Prompt Strategies}
\label{tab:performance}
\resizebox{\textwidth}{!}{%
\begin{tabular}{l*{10}{c}}
\toprule
\multirow{3}{*}{\makecell[l]{Model +\\ Prompt Strategy}} & 
\multicolumn{4}{c}{Conflict Detection} & 
\multicolumn{2}{c}{Type Detection} & 
\multicolumn{4}{c}{Segmentation} \\
\cmidrule(lr){2-5} \cmidrule(lr){6-7} \cmidrule(lr){8-11}
& & & & & & & \multicolumn{2}{c}{Guided} & \multicolumn{2}{c}{Blind} \\
\cmidrule(lr){8-9} \cmidrule(lr){10-11}
& Accuracy & Precision & Recall & F1 & Accuracy & Macro F1 & Jaccard & F1 & Jaccard & F1 \\
\midrule
Claude-3 Sonnet + Basic & 0.539 & 0.901 & 0.296 & 0.446 & \textbf{0.401} & \textbf{0.216} & 0.582 & 0.601 & 0.562 & 0.538 \\
Claude-3 Sonnet + CoT   & \textbf{0.710} & 0.951 & \textbf{0.566} & \textbf{0.710} & 0.368 & 0.119 & 0.551 & 0.586 & \textbf{0.624} & 0.602  \\
Claude-3 Haiku + Basic  & 0.395 & 0.913 & 0.036 & 0.069 & 0.278 & 0.174 & 0.521 & 0.545 & 0.577  & 0.596 \\
Claude-3 Haiku + CoT    & 0.578 & 0.948 & 0.344 & 0.505 & 0.282 & 0.135 & 0.573 & 0.598 & 0.500 & \textbf{0.606} \\
Llama3.3-70B + Basic       & 0.679 & 0.916 & 0.535 & 0.676 & 0.308 & 0.065 & \textbf{0.727} & \textbf{0.734} & 0.547 &  0.587 \\
Llama3.3-70B + CoT         & 0.497 & \textbf{0.987} & 0.198 & 0.331 & 0.245 & 0.095 & 0.712 & 0.726 & 0.541 &  0.577 \\
Llama3.1-8B + Basic        & 0.380 & 0.812 & 0.01 & 0.02 & 0.328 & 0.163 & 0.395 & 0.436 & 0.353  & 0.385 \\
Llama3.1-8B + CoT          & 0.482 & 0.699 & 0.301 & 0.421 & 0.399 & 0.056 & 0.397 & 0.430 & 0.336  & 0.373 \\
\bottomrule
\end{tabular}%
}
\end{table*}
\subsection{Type of Conflict}
In this task, we give a set of documents with a contradiction and ask the model to predict what type of conflict exists in the documents: self-contradictions, pair contradictions or conditional contradictions. The objective of the task is to analyze how well models can understand the nuances of contradictions. 
\subsection{Conflicting Context Segmentation}
The objective of this task is to identify which document(s) contain conflicting information within a given set. We design two variants of this task, "Guided Segmentation" , requires the model to identify the conflicting documents when provided with the type of conflict present in the set. This task evaluates the model's ability to leverage known conflict types in pinpointing contradictions. The second, more challenging variant, called "Blind Segmentation", tasks the model with correctly identifying contradictory documents without prior knowledge of the conflict type.  To evaluate performance on these tasks, we frame them as multi-label classification problems. We employ two metrics: Jaccard similarity and F1 score to evaluate the performance of LLMs.

\subsection{Model Selection and Prompting Strategies}
We experiment with both different model architectures and prompting approaches. We employ four state-of-the-art LLMs that represent a range of model sizes. Among the larger models, we use Claude-3 Sonnet \cite{Anthropic} and Llama-3.3 70B \cite{touvron2023llama}. For smaller-scale models, we evaluate Claude-3 Haiku \cite{Anthropic}, a more efficient variant of Claude-3, and Llama-3.1 8B, a lightweight version of Llama. This allows us to understand both the impact of model scale (70B vs 8B parameters) and architectural differences (Claude vs Llama) on contradiction detection performance.

For each model, we investigate two prompting strategies. Basic prompting provides direct instructions that explicitly state the task requirements without additional guidance or structure. Chain-of-Thought (CoT) prompting encourages step-by-step reasoning by breaking down the contradiction detection process into logical steps, following the methodology proposed by \cite{wei2022chain}.

\section{Results}
In the task of \textbf{conflict detection}, Claude-3 Sonnet with CoT prompting outperforms other models. The impact of prompting strategy is mixed for different model families: while CoT improves Claude models' performance (31\% increase for Sonnet, 46\% for Haiku), it degrades Llama models' performance (26\% decrease for Llama-70B). Regarding model size, larger variants (Claude-3 sonnet, Llama-70b) outperform their smaller counterparts under the same prompting strategy.

We observe that all models demonstrate high precision but lower recall, suggesting that models are highly conservative in their contradiction predictions. \textbf{This indicates that while models are very reliable when they do flag a contradiction, they miss many actual contradictions.}
 
In \textbf{type detection}, Claude-3 Sonnet with basic prompting achieves the highest performance. Contrary to expectations, CoT prompting decreases performance across most models, with performance drops ranging from 8\% for Claude models up to 25\% for Llama 70B. The size of the model has mixed effects - while Claude-3 Sonnet outperforms Haiku, the smaller Llama-8B outperforms the larger 70B variant by about 6\%, suggesting that \textbf{type detection may rely more on the model's fundamental understanding of contradictions rather than raw computational power or reasoning prompts.}

\begin{figure*}[t]
    \centering
    \begin{subfigure}[b]{0.48\textwidth}
        \centering
        \includegraphics[width=\textwidth]{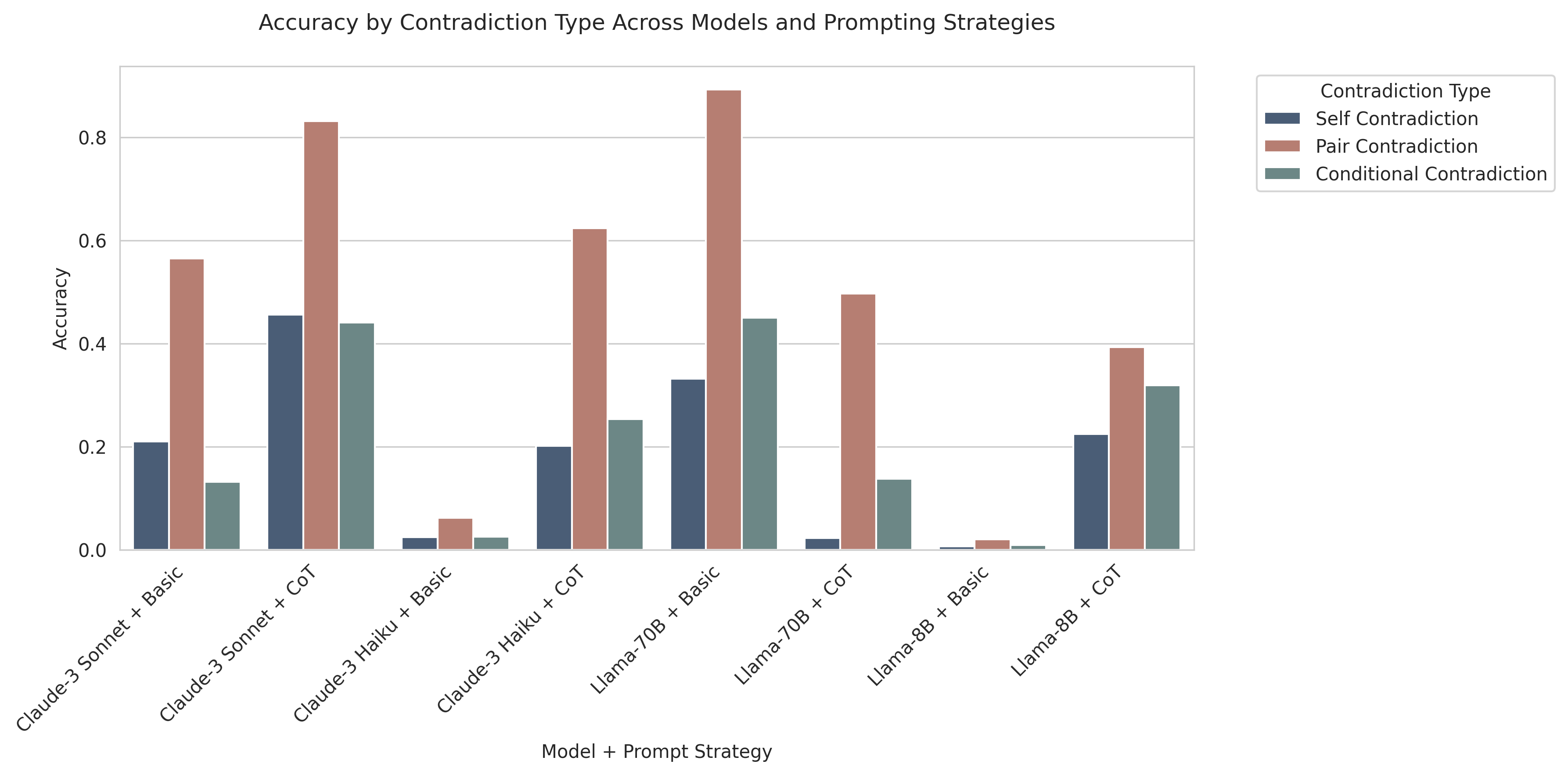}
        \caption{Performance across different types of contradictions}
        \label{fig:contradiction_types}
    \end{subfigure}
    \hfill
    \begin{subfigure}[b]{0.48\textwidth}
        \centering
        \includegraphics[width=\textwidth]{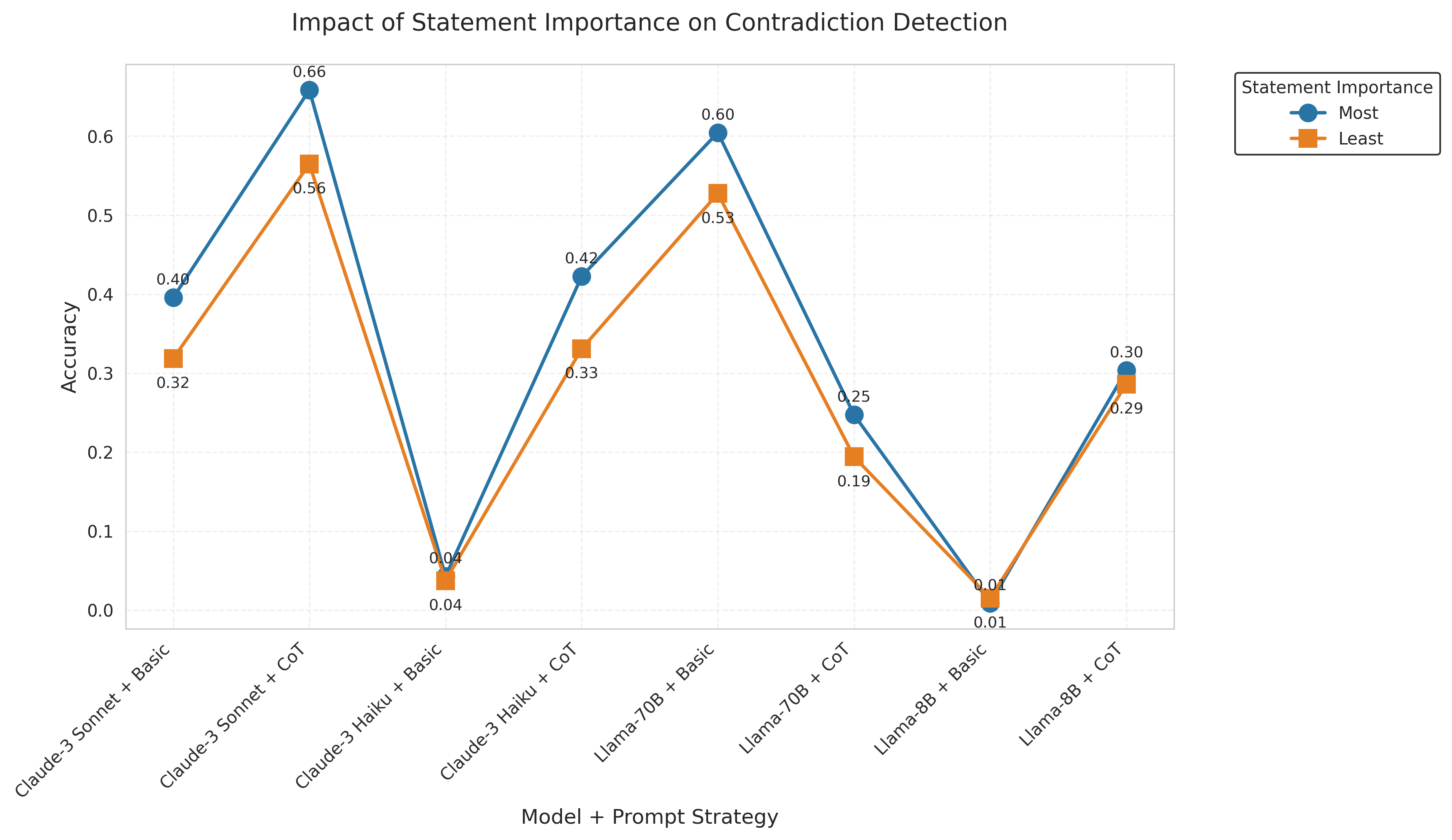}
        \caption{Impact of statement importance on detection}
        \label{fig:importance}
    \end{subfigure}
    \caption{Analysis of contradiction detection performance: (a) comparison across different contradiction types (self, pair, and conditional) and (b) effect of statement importance (most vs. least) on detection accuracy across different models and prompting strategies.}
    \label{fig:combined}
\end{figure*}
The \textbf{segmentation results} reveal interesting patterns across both guided and blind scenarios. Llama-70B with basic prompting achieves the best performance in guided segmentation. However, in blind segmentation, Claude-3 Sonnet with CoT shows superior performance. There is a varied impact of CoT promprting strategy with 1-2\% degradation in performance for Llama models but no clear improvement / degradation for Claude models across the 2 segmentation tasks. This suggests that, the \textbf{effectiveness of prompting strategies in complex tasks such as segmentation is highly model-dependent, and that larger models generally have an advantage}. The consistently higher scores in guided versus blind segmentation across most models indicates the value of providing type information for accurate contradiction localization.

\section{Ablation Studies}
\begin{figure*}[t]
    \centering
    \begin{subfigure}{0.48\linewidth}
        \centering
        \includegraphics[width=\linewidth]{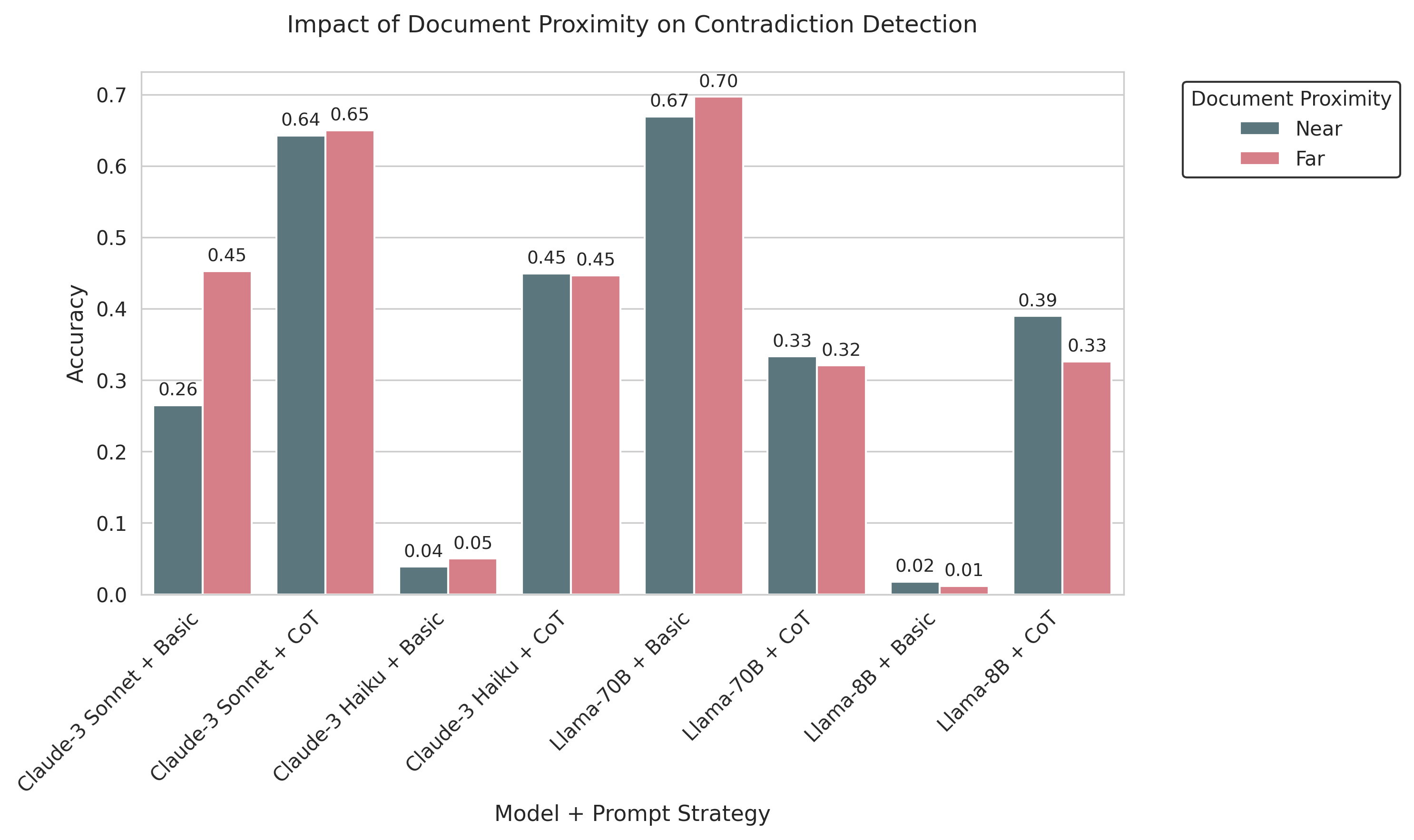}
        \caption{Impact of document proximity}
        \label{fig:proximity}
    \end{subfigure}
    \hfill
    \begin{subfigure}{0.48\linewidth}
        \centering
        \includegraphics[width=\linewidth]{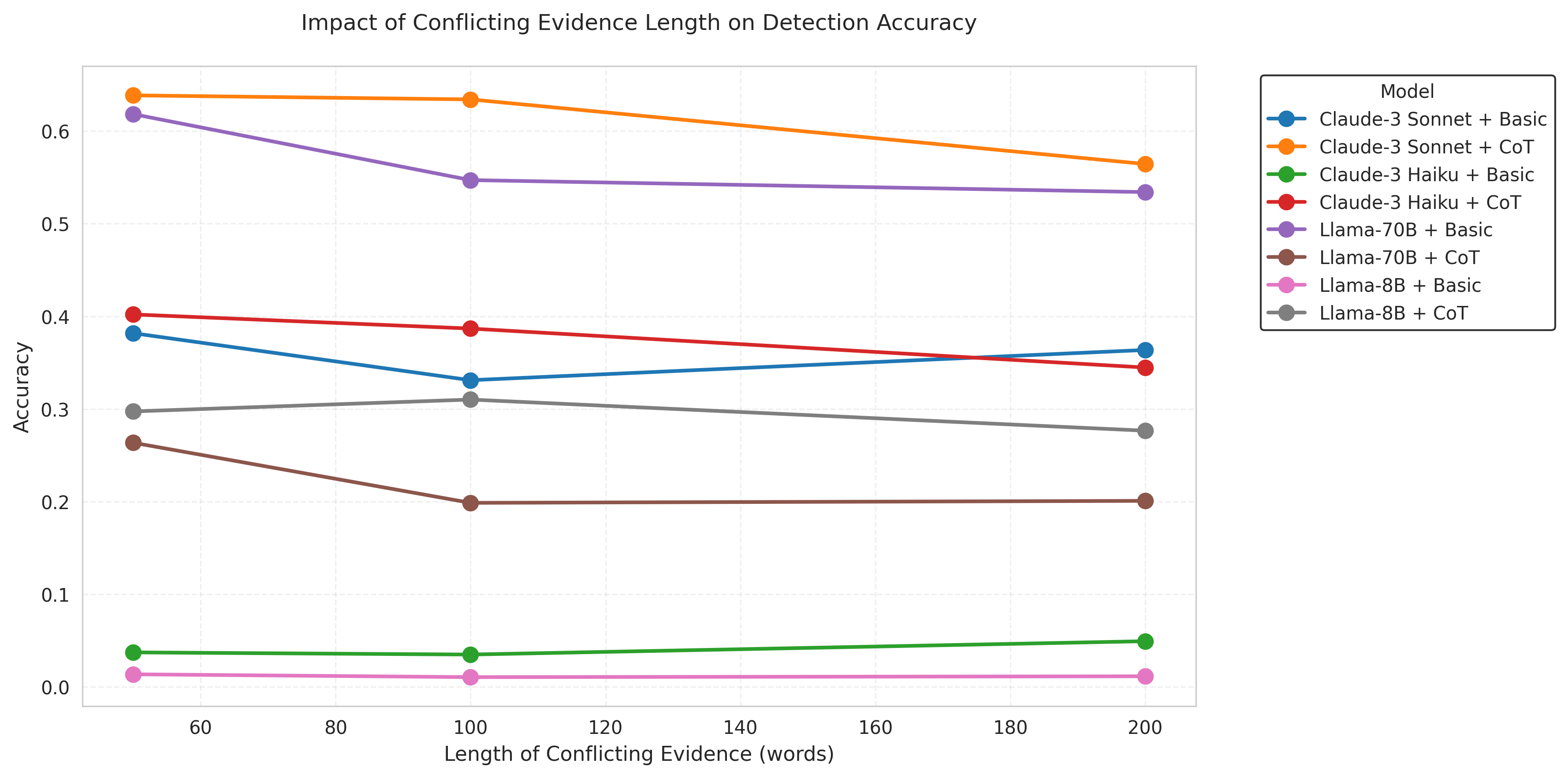}
        \caption{Impact of conflicting evidence length}
        \label{fig:evidence_length}
    \end{subfigure}
    \caption{Analysis of positioning and evidence length effects: (a) performance comparison between near and far document positioning, and (b) impact of conflicting evidence length on detection accuracy across different models and prompting strategies.}
    \label{fig:position_length_analysis}
\end{figure*}

\textbf{RQ1: How does the type of contradiction (self, pair, or conditional) affect the model's detection accuracy?}
There are notable differences in model performance across different types of contradictions (see Figure 2a). Pair contradictions are consistently easier to detect across all models and prompting strategies, with accuracy rates substantially higher than other contradiction types. For instance, Llama-70B with basic prompting achieves its highest accuracy of 0.893 on pair contradictions, while Claude-3 Sonnet with CoT reaches 0.831 for the same type. Conditional contradictions and self-contradictions prove to be more challenging to detect, with generally lower accuracy rates across all models. Self-contradictions show particularly low detection rates, with accuracies ranging from 0.006 to 0.456, suggesting that identifying contradictions within a single document is difficult for LLMs. The relative difficulty of contradiction types follows a consistent pattern across models: pair contradictions are the easiest to detect, followed by conditional contradictions, while self-contradictions are generally the most challenging. \textbf{This hierarchy suggests that LLMs are better equipped to compare and contrast information across distinct documents than to analyze internal consistency within a single document or understand complex conditional relationships.}

\textbf{RQ2: To what extent does the importance of conflicting statements influence the model's ability to detect contradictions?}
The analysis of statement importance (Refer to Sec.~\ref{sec:llmcv} for definition) reveals a consistent pattern (Figure 2b) across most models, with important statements generally leading to better contradiction detection. All models except Llama-8B Basic show improved performance when dealing with more important statements. Larger models appear to be more sensitive to statement importance, as evidenced by the substantial differences observed in Claude-3 and Llama-70B models compared to Llama-8B. Chain-of-thought (CoT) prompting appears to amplify the importance effect in Claude models, with both Sonnet and Haiku variants showing larger performance gaps  compared to their basic prompting counterparts . The consistent impact of statement importance across most models suggests that \textbf{LLMs are inherently better at identifying contradictions in semantically significant statements.} However, the magnitude of this effect varies considerably, from negligible in simpler models to substantial in more advanced architectures. 

\textbf{RQ3: How do the relative positions of conflicting documents within the input set impact the model's performance in identifying contradictions?}
Most models show comparable or slightly better performance when contradicting documents are positioned far apart rather than near each other (Figure 3a). This is particularly evident in Claude-3 Sonnet with basic prompting, which shows a substantial 18.8 percentage point improvement in accuracy when documents are positioned far apart. Larger models (Claude-3 Sonnet and Llama-70B) generally maintain more consistent performance across different document positions. CoT prompting seems to stabilize performance across positions, as evidenced by the smaller positional impact compared to basic prompting. For instance, Claude-3 Sonnet's position sensitivity decreases dramatically from 18.8 points with basic prompting to just 0.7 points with CoT. The results suggest that sophisticated models, particularly when enhanced with \textbf{CoT prompting, can effectively identify contradictions regardless of document proximity}.

\textbf{RQ4: What is the relationship between the amount of conflicting information and the model's detection accuracy?}
Most models show a slight decline in performance as the length of conflicting evidence increases, suggesting that longer conflicting segments may make contradiction detection more challenging. This trend is most pronounced in larger models, with Llama-70B Basic showing a decrease from 61.8\% accuracy for short evidence (1-50 words) to 53.4\% for longer evidence (151-200 words). 

\section{Conclusion and Future Work}
In this work, we introduced a framework for generating and evaluating different types of contradictions. Our experiments with various LLMs and prompting strategies revealed both the capabilities and limitations of current models in serving as context validators. In the future, we would like to experiment with more robust quality control mechanims to ensure the quality of the synthetic data. We would also like to experiemnt with more types and sub-types of conflicts such as numerical inconsistencies, temporal contradictions etc. Finally, an important direction for future work is developing methods to resolve detected contradictions. This includes not only identifying conflicts but also determining which information is more reliable. Strategies for conflict resolution could range from simple heuristics based on document metadata to more sophisticated approaches that consider source credibility, temporal relationships, and logical consistency. Understanding how to effectively present and resolve contradictions to end users is also crucial for building trustworthy RAG systems.

\section{Limitations}
In this work we focus on three types of contradictions in retrieved documents. It might be possible that there are more sub-categories or categories of conflicts that occur in real world RAG systems like numerical, logical, temporal  or causal conflicts etc. Additionally, our proposed framework does not have a quality control mechanism and is dependent on human annotation, limiting its scalability. We have limited our experiments to use LLMs such as Claude 3 and Llama. Models such as GPT-4 might follow a different pattern compared to our findings. 
\bibliography{acl_latex}
\newpage
\section*{Appendix}


\subsection*{A.1 Prompts for Data Generation}
\begin{tcolorbox}[title=ChooseStatement Prompt,
    colback=white,
    colframe=gray,
    width=\columnwidth]
Choose the {importance} important sentence from the given document. 
Only output the sentences within <sentence></sentence> tags.
Here is the document: {document}
\end{tcolorbox}
where \texttt{importance} can be either ``most'' or ``least''.

\begin{tcolorbox}[title=ContradictStatement Prompt,
    colback=white,
    colframe=gray,
    width=\columnwidth]
Modify the given statement to suggest otherwise instead of the original. 
Only output the modified statement within <statement></statement> tags.
Here is the statement: {statement}
\end{tcolorbox}

\begin{tcolorbox}[title=ContextGenerate Prompt,
    colback=white,
    colframe=gray,
    width=\columnwidth]
Generate a paragraph of {length} words continuing the given sentence. 
Only output the paragraph within <paragraph></paragraph> tags.
Here is the sentence: {sentence}
\end{tcolorbox}
where \texttt{length} specifies the desired word count for the generated context.

\begin{tcolorbox}[title=GenerateConditionalContradiction Prompt,
    colback=white,
    colframe=gray,
    width=\columnwidth]
Generate a set of three short documents about a the given topic. Follow these rules:
Document 1 and Document 2 should provide different, non-contradictory information 
about the same topic. Document 1 and 2 should not contradict each other.
Information in Document 3 should not contradict information in Document 1.
Information in Document 3 should not contradict information in Document 2.
The information in Document 3 should create a conditional contradiction between Document 1 
and Document 2, making them mutually exclusive given the context provided in Document 3. 
This means that while Documents 1 and 2 can both be true in isolation, 
they cannot both be true when the information in Document 3 is considered. 
Make sure document 3 sounds realistic. 
Format the output as follows: 
<document1> [Content of Document 1] </document1> 
<document2> [Content of Document 2] </document2> 
<document3> [Content of Document 3] </document3> 
Ensure that each document is concise, clear, and focused on a single aspect of the topic. 
The conditional contradiction should emerge naturally from the combination of 
all three documents, making it impossible for both Document 1 and Document 2 
to be true simultaneously when Document 3 is taken into account.
Here is an example:
<document1>:
The Smith family always vacations in tropical locations during winter.
</document1>
<document2>
The Smiths enjoy skiing and snowboarding every winter.
</document2>
<document3>
The Smith family has a strict policy of taking only one vacation per year, 
which they always schedule during the winter months.
</document3>
Here is the topic: {firstsentence}
\end{tcolorbox}
where \texttt{firstsentence} is the first sentence of the sampled document.

\subsection*{A.3 Prompts for Context Validator}
This section lists all the prompts used for the three tasks. The prompts shown here are for the CoT prompting strategy. For basic strategy, we remove instruction "Think Step by Step".
\begin{tcolorbox}[title=Conflict Detection Prompt,
    colback=white,
    colframe=gray,
    width=\columnwidth]
You are given a set of documents. Do the documents contain conflicting information? Answer yes or no. Think step by step before answering.
\end{tcolorbox}

\begin{tcolorbox}[title=Conflict Type Prediction Prompt,
    colback=white,
    colframe=gray,
    width=\columnwidth]
Given a set of documents with a contradiction, your task is to predict the type of
contradiction present, if any. The possible types are:

1. Self-Contradiction: Conflicting information within a single document.
2. Pair Contradiction: Conflicting information between two documents.
3. Conditional Contradiction: Three documents where the third document makes the first two
contradict each other.

Instructions:
1. Carefully read all the provided documents.
2. Analyze the content for any contradictions within or between documents.
3. Determine the type of contradiction based on the definitions provided.
4. Return the type of contradiction within <type> </type> tags.
5. Think step by step before answering.
\end{tcolorbox}

\begin{tcolorbox}[title=Guided Segmentation,
    colback=white,
    colframe=gray,
    width=\columnwidth]
Given a set of documents and a known conflict type, your task is to identify 
which document(s) id contain the conflicting information.

Conflict Type: {conflict type}
Instructions:
1. Carefully read all the provided documents.
2. Keep in mind the given conflict type.
3. Analyze the content to identify which document(s) contribute to the specified 
type of contradiction.
4. List the numbers of the documents that contain the conflicting information.
5. Think step by step before answering.

Your response should be in the following format:
<documents>[List the numbers of the documents, separated by commas]</documents>

Definitions of Conflict Types:
- Self-Contradiction: Conflicting information within a single document.
- Pair Contradiction: Conflicting information between two documents.
- Conditional Contradiction: Three documents where the third document makes the first two 
contradict each other, although they don't contradict directly.

Here are the documents: {documents}
\end{tcolorbox}

\begin{tcolorbox}[title=Blind Segmentation,
    colback=white,
    colframe=gray,
    width=\columnwidth]
Given a set of documents, your task is to identify which document(s) id contain the 
conflicting information.
Instructions:
1. Carefully read all the provided documents.
2. Analyze the content to identify which document(s) contribute to the 
specified type of contradiction.
3. List the numbers of the documents that contain the conflicting information.
4. Think step by step before answering.

Your response should be in the following format:
<documents>[List the numbers of the documents, separated by commas]</documents>
Here are the documents: {documents}
\end{tcolorbox}
\begin{table*}[t]
\centering
\caption{Sensitivity Analysis: Conflict Detection performance across 2 runs}
\begin{tabular}{lrr}
\toprule 
 & Pair Contradiction & Self Contradiction \\
Model  &  &  \\
\midrule
Claude-3 Haiku + Basic & $0.057 \pm 0.0045$ & $0.023 \pm 0.001$ \\
Claude-3 Haiku + CoT & $0.640 \pm 0.016$ & $0.204 \pm 0.001$ \\
Claude-3 Sonnet + Basic & $0.567 \pm 0.002$ & $0.210 \pm 0.00$ \\
Claude-3 Sonnet + CoT & $0.831 \pm 0.00$ & $0.454 \pm 0.002$ \\
Llama-70B + Basic & $0.890 \pm 0.002$ & $0.333 \pm 0.001$ \\
Llama-70B + CoT & $0.488 \pm 0.008$ & $0.025 \pm 0.003$ \\
Llama-8B + Basic & $0.020 \pm 0.00$ & $0.007 \pm 0.001$ \\
Llama-8B + CoT & $0.393 \pm 0.00$ & $0.225 \pm 0.001$ \\
\bottomrule
\end{tabular}
\end{table*}

\subsection*{A.4 Annotator Instructions}
\textbf{Self Contradictions:}
Analyze the given document for contradictions. Answer yes/no, if the document has information that is contradicting with itself. 

\textbf{Pair Contradictions:}
Analyze the given pair of documents. Answer yes/no, if the information in two documents are contradicting each other.

\textbf{Conditional Contradictions:}
Analyze the given set of 3 documents. The set of documents are conditionally conflicting if the following rules are satisfied:
\begin{itemize}
    \item Document 1 and 2 do not contradict each other
    \item Document 3 makes document 1 and 2 contradict/ not true together
\end{itemize}

\begin{table*}[h]
\centering
\begin{tabular}{|p{2.5cm}|p{12cm}|}
\hline
\textbf{Conflict Type} & \textbf{Example} \\
\hline
Self-contradiction & \textbf{Document:} "Calvin Tyler Scott is a Canadian basketball player for the UPEI Panthers. Tyler Scott was born and raised in Halifax, Nova Scotia. Tyler Scott attended Halifax West High School and was the top scorer for the Halifax West Warriors. After graduating from Halifax West, Tyler Scott attended Lee Academy, a prep school in Maine. After Lee Academy, Tyler Scott went to Acadia University in New Minas, where he averaged 11.7 points per game, after realizing Acadia wasn't where he felt 100\% comfortable he committed to UPEI with Tim Kendrick. At UPEI Tyler Scott went on to average 23 points per game in his first year and became a first team all Canadian and during his second and third year at UPEI, Tyler Scott was named second team all star and was 2nd in scoring in the AUS and 1st in scoring in his 5th year. On February 26, 2017, Tyler Scott made it into top 5 AUS scoring of all time. On February 26, 2017, Tyler Scott did not make it into the top 5 AUS scoring of all time.Despite not achieving the coveted top 5 AUS scoring record, Tyler Scott's performance on that fateful day in February 2017 was nothing short of remarkable. With unwavering determination and a relentless drive to excel, he pushed himself to the limits, leaving everything on the court. While the elusive record may have eluded him, his efforts served as an inspiration to his teammates and fans alike. Tyler's journey was a testament to the power of perseverance, reminding everyone that true greatness lies not in the accolades achieved but in the pursuit of excellence itself. His legacy transcended mere statistics, etching his name in the annals of AUS history as a true champion of the game.  During his 5th year Tyler Scott also passed 1700 career points." \\
\hline
Pair contradiction & \textbf{Document 1:} "Reynolds v. United States, 98 U.S. (8 Otto.) 145 (1878), was a Supreme Court of the United States case that held that religious duty was not a defense to a criminal indictment. "Reynolds" was the first Supreme Court opinion to address the Impartial Jury and the Confrontation Clauses of the Sixth Amendment."
\newline \textbf{Document 2:} "Reynolds v. United States, 98 U.S. (8 Otto.) 145 (1878), was a Supreme Court of the United States case that upheld religious duty as a valid defense to a criminal indictment.The Court ruled that a member of a religious group that prohibited work on Sundays could not be prosecuted for violating a federal law prohibiting labor on Sundays. This decision established the principle that the government cannot compel individuals to violate their religious beliefs, setting an important precedent for the protection of religious freedom in the United States." \\
\hline
\end{tabular}
\caption{Examples where annotators marked documents as not conflicting, but they are conflicting.}
\label{tab:conflicts}
\end{table*}
\newpage


\end{document}